\documentclass[11pt]{article}

\usepackage{acl}

\usepackage{times}
\usepackage{latexsym}

\usepackage[T1]{fontenc}

\usepackage[utf8]{inputenc}

\usepackage{microtype}

\usepackage{inconsolata}

\usepackage{graphicx}

\title{PERCEPT: A Corpus for POS Tagging and Analysis of Persian--English Code-Mixing}

\author{
  Ghazal Kalhor\textsuperscript{1} \quad
  Zahra Jafari\textsuperscript{2}\thanks{Equal contribution.} \quad
  Amirarsalan Shahbazi\textsuperscript{2}\footnotemark[1] \quad
  Behnam Bahrak\textsuperscript{3}
  \\
  \textsuperscript{1}EPFL \quad
  \textsuperscript{2}University of Tehran \quad
  \textsuperscript{3}Khatam University
  \\
  \small{
    \textbf{Correspondence:}
    \href{mailto:ghazal.kalhor@epfl.ch}{ghazal.kalhor@epfl.ch},
    \href{mailto:b.bahrak@teias.institute}{b.bahrak@teias.institute}
  }
}

\usepackage{booktabs}
\usepackage{xltabular}
\usepackage{array}
\usepackage{caption}

\begin{document}
\maketitle

\begin{abstract}
Social media has become a major venue for multilingual communication, where users frequently mix multiple languages within a single utterance. Although code-mixed corpora have been developed for several language pairs, Persian--English code-mixing remains relatively underexplored. Existing Persian resources lack Universal Dependencies (UD) part-of-speech (POS) annotations for code-mixed words, limiting both linguistic analyses and the development of syntax-aware NLP models. To address this gap, we introduce \textsc{PERCEPT}, the first publicly available large-scale Persian--English code-mixed corpus annotated with Universal Dependencies POS tags for code-mixed words. The dataset comprises 6,800 posts collected from Twitter (X), Instagram, and Digikala, each containing at least one identified Persian--English code-mixed word. We further present an LLM-assisted annotation framework that automatically assigns POS tags and document-level topics. Human evaluation demonstrates high agreement between the automatically generated annotations and gold annotations, confirming the reliability of the annotations. Using \textsc{PERCEPT}, we conduct the first comprehensive linguistic analysis of Persian--English code-mixing across multiple social media platforms. Our analyses reveal that nouns are the predominant category for code-mixed words, while the distributions of other POS categories vary across platforms. We further find that the positional distribution of code-mixed words is remarkably consistent across platforms, whereas the triggering effect is substantially more pronounced in Digikala. \textsc{PERCEPT} is publicly available at \url{https://github.com/kalhorghazal/PERCEPT}.
\end{abstract}

\begin{figure}[!t]
  \centering
  \includegraphics[width=\columnwidth]{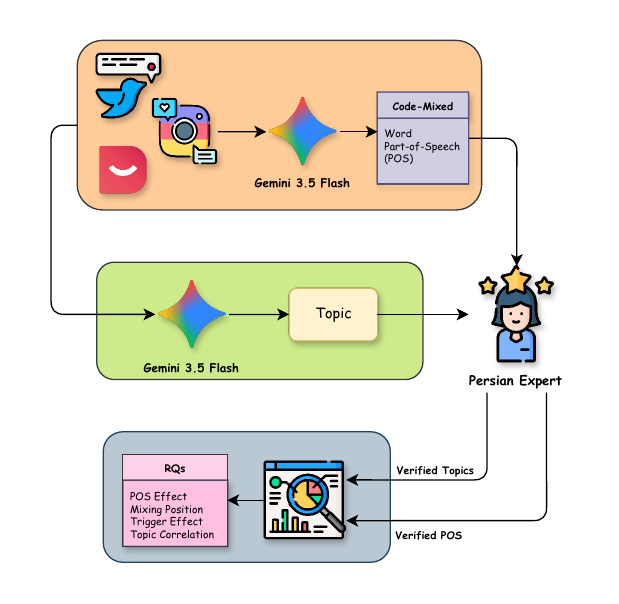}
  \caption{Pipeline illustrating the steps for data annotation and statistical analysis of Persian--English code-mixed text.}
  \label{fig:pipeline}
\end{figure}

\section{Introduction}

Social media has become one of the primary settings in which multilingual speakers communicate. In these environments, users frequently mix elements from multiple languages within the same utterance, a phenomenon commonly referred to as \textit{code-mixing} \cite{nordin2023code}. Persian-speaking users are no exception, often incorporating English words and expressions into otherwise Persian posts on platforms such as Twitter (X), Instagram, and Digikala. Understanding how code-mixing is realized in Persian is important not only for linguistic research but also for the development of robust natural language processing (NLP) systems. The presence of multiple languages, transliterated words, and inconsistent spelling introduces additional challenges for tasks such as tokenization, language identification, part-of-speech (POS) tagging, syntactic parsing, and sentiment analysis \cite{vyas2014pos, barman2019automatic, singh2018automatic}.

Despite the growing body of work on code-mixed NLP, Persian--English code-mixing remains comparatively underexplored. Although code-mixed corpora have been introduced for several language pairs, including Malayalam--English \cite{chakravarthi2020sentiment}, Bengali--English \cite{alam2025bnsentmix}, Hindi--English \cite{kumar2018aggression}, and Sinhala--English \cite{uthpala2024sinhala}, comparable resources for Persian remain scarce. Existing Persian datasets primarily target language identification \cite{ghafouri2025pinlid} or sentiment analysis \cite{sabri2021sentiment} and do not provide POS annotations for code-mixed words, limiting both linguistic analyses and the development of syntax-aware NLP models.

To address this gap, we introduce \textsc{PERCEPT}, a dataset of 6,800 Persian--English code-mixed texts collected from three major social media and e-commerce platforms: Twitter (X), Instagram, and Digikala. By construction, each text contains at least one identified Persian--English code-mixed word. The dataset contains code-mixed words written both in English and transliterated into Persian. Using Gemini 3.5 Flash \cite{google2026gemini35}, a state-of-the-art large language model (LLM) in the Gemini family that has shown strong performance on Persian-language tasks,\footnote{\url{https://mcinext-mizan-llm-leaderboard.hf.space/}} we automatically identify the POS of each code-mixed word according to the Universal Dependencies tagset and assign a topic label to every text. Human validation demonstrates high agreement between the automatically generated annotations and gold annotations, confirming the reliability of the resulting dataset.

Beyond introducing a new resource, \textsc{PERCEPT} enables a comprehensive empirical analysis of Persian--English code-mixing. Specifically, we investigate four questions: (1) which POS categories are most likely to host code-mixing; (2) where code-mixed words tend to occur within an utterance; (3) whether code-mixing exhibits a triggering effect by recurring within the same text; and (4) how the degree of code-mixing varies across discussion topics.

Our analyses reveal that code-mixing in Persian social media is predominantly realized through nouns, while the distributions of other POS categories vary across platforms. We further find that the positional distribution of code-mixed words is remarkably consistent across Twitter (X), Instagram, and Digikala, whereas the triggering effect is substantially more pronounced in Digikala. Finally, our topical analysis shows that texts related to \textit{Industry \& Commerce} and \textit{Brand \& Business} exhibit the highest degree of code-mixing.

The main contributions of this work are as follows:

\begin{itemize}
\item We introduce \textsc{PERCEPT}, the first large-scale Persian--English code-mixed corpus annotated with Universal Dependencies POS tags for code-mixed words and document-level topic labels.
\item We propose an LLM-assisted annotation framework for constructing linguistically annotated code-mixed corpora and demonstrate its reliability through human validation.
\item We present the first large-scale empirical analysis of Persian--English code-mixing across multiple social media platforms, providing insights into POS distributions, positional preferences, triggering effects, and topical variation.
\end{itemize}

\section{Related Work}

\paragraph{Code-Mixed Corpora}

Several code-mixed corpora have been introduced for language pairs in which English serves as the embedded language. \citet{chakravarthi2020sentiment} propose a dataset of 6,739 Malayalam--English code-mixed YouTube comments annotated for sentiment analysis. Similarly, \citet{alam2025bnsentmix} introduce a dataset of 20,000 Bengali--English code-mixed texts collected from Facebook, YouTube, and e-commerce platforms, while \citet{kumar2018aggression} compile a Hindi--English code-mixed dataset from Facebook and Twitter for aggression detection. \citet{uthpala2024sinhala} also create a Sinhala--English code-mixed dataset from YouTube comments for sentiment analysis. Although these resources have substantially advanced research on code-mixed NLP, most focus on downstream applications such as sentiment analysis, whereas linguistically annotated corpora supporting tasks such as POS tagging remain relatively scarce.

\paragraph{Code-Mixed POS Tagging}

Several studies have introduced POS-annotated datasets for code-mixed text across different language pairs. \citet{solorio2008part} investigate POS tagging of English--Spanish code-switched text, while \citet{vyas2014pos} study POS tagging of English--Hindi social media content. \citet{barman2016part} explore different approaches, including pipeline, stacking, and joint modeling, for POS tagging of trilingual English--Bengali--Hindi code-mixed data collected from Facebook posts and comments. Similarly, \citet{singh2018twitter} introduce a POS-tagged dataset of 1,489 Hindi--English code-mixed tweets, and \citet{jamatia2016collecting} investigate POS annotation of English--Bengali and English--Hindi social media text. Beyond social media, \citet{bhat2018universal} present a Hindi--English code-switching treebank annotated under Universal Dependencies, while \citet{ccetinouglu2023two} introduce a Turkish--German treebank with POS tags, morphological features, and dependency annotations. More recently, \citet{sheth2025comi} introduce COMI-LINGUA, a large Hindi--English dataset supporting multiple NLP tasks, including POS tagging and named entity recognition. Despite these advances, to the best of our knowledge, no publicly available large-scale dataset specifically designed for Persian--English code-mixed text provides POS annotations for code-mixed words.

\paragraph{Persian Resources}

Only a limited number of studies have investigated Persian--English code-mixed text and speech. \citet{sabri2021sentiment} collect 3,640 tweets by searching for a list of 44 English words transliterated into Persian and manually annotate the data for sentiment analysis. In another study, \citet{moradi2019structural} investigate the structural characteristics of Persian--English code-mixing and code-switching through interviews with bilingual Iranian students. More recently, \citet{ghafouri2025pinlid} introduce PinLID, a dataset of 6,915 Persian--English code-mixed tweets collected by searching for English words in Persian tweets. However, the dataset does not consider English words transliterated into Persian, and its analysis is limited to sentence- and token-level language identification. In contrast, \textsc{PERCEPT} provides the first large-scale POS-annotated Persian--English code-mixed corpus, covering both English-script and transliterated English words, thereby enabling fine-grained linguistic analyses beyond language identification.

\section{PERCEPT Dataset}

To construct our dataset, we consider three major social media and e-commerce platforms: Twitter (X), Instagram, and Digikala.\footnote{\url{https://www.digikala.com/}} These platforms cover diverse topics and are widely used by Persian-speaking users, who often communicate in a conversational style. We identify prior studies that collect and publish Persian-language data from these platforms. Table~\ref{tab:dataStat} summarizes the source datasets, including the total number of texts analyzed in the original studies and the number retained for \textsc{PERCEPT}. The datasets incorporated into \textsc{PERCEPT} originate from prior academic research and are accessed through the mechanisms established by their respective authors. Some are publicly available through links provided with the original publications, while others are obtained directly from the corresponding authors upon request. We use these resources for POS annotation and analysis of Persian--English code-mixing, while retaining the access and usage constraints associated with the original datasets. To remove offensive content, we retain only instances labeled as \textit{clean} in datasets with offensive-language annotations and manually filter offensive content from the remaining datasets. We then filter the source datasets for Persian--English code-mixed content, retaining 5,627 texts: 2,845 tweets from Twitter (X), 1,463 comments from Instagram, and 1,319 comments from Digikala, out of 180,136 source texts.

\begin{table*}[h]
\centering
\small
\begin{tabular}{llll}
\toprule
\textbf{Data Source} & \textbf{Platform} & \textbf{No. Analyzed} & \textbf{No. in PERCEPT}\\
\midrule
\citet{golazizian2020irony} & Twitter (X) & 2,942 & 315 \\
\citet{sabri2021emopars} & Twitter (X) & 30,000 & 1,578 \\
\citet{kebriaei2024persian} & Twitter (X) & 18,628 & 695 \\
\citet{elahimanesh2025emotion} & Twitter (X) & 3,352 & 257 \\
\citet{heidari2020producing} & Instagram & 95,749 & 1,169 \\
\citet{nazarizadeh2025parsoffensive} & Instagram & 5,214 & 81 \\
\citet{davardoust2024dark} & Instagram & 15,000 & 213 \\
\citet{asli2020optimizing} & Digikala & 10,000 & 1,319 \\
\bottomrule
\end{tabular}
\caption{Statistics of datasets used to construct PERCEPT, including source, platform, and number of analyzed and retained instances.}
\label{tab:dataStat}
\end{table*}

To identify additional code-mixed content on Twitter (X), we first apply the POS annotation prompts, described in Section~\ref{sec:posAnnotate}, to the 2,845 retained tweets. Based on these annotations, we identify the 71 most frequent Persian--English code-mixed words written in Persian script. We then manually search Twitter (X) for additional tweets containing these words, aiming to collect at least 20 tweets per word whenever available. For each word, we enter a query of the form \texttt{("code-mixed word") lang:fa -is:retweet -is:reply since:2026-01-01} directly into the Twitter (X) search bar. The \texttt{lang:fa}, \texttt{-is:retweet}, and \texttt{-is:reply} operators restrict the results to Persian-language tweets while excluding retweets and replies; \texttt{since:2026-01-01} restricts results to tweets posted on or after January 1, 2026. Since the source tweets are collected before 2026, this date restriction helps avoid overlap with the existing dataset. After removing offensive content, we obtain 1,173 additional tweets.

The final dataset comprises 6,800 texts containing Persian--English code-mixed words. Of these, 5,627 (82.8\%) come from previously collected datasets and 1,173 (17.2\%) from targeted searches. Within the Twitter (X) subset, 2,845 texts (70.8\%) come from previously collected datasets and 1,173 (29.2\%) from targeted searches; the Instagram and Digikala subsets consist entirely of previously collected texts. Before annotation, we anonymize all texts by replacing user mentions with \texttt{@username} and URLs with \texttt{www.link.com}. Figure~\ref{fig:pipeline} summarizes the dataset construction and analysis pipeline. We next describe each annotation stage in detail.

\subsection{POS Annotation}\label{sec:posAnnotate}

To assign part-of-speech tags to code-mixed words in tweets and comments, we query Gemini 3.5 Flash, providing it with batches of 50 or 100 texts and asking it to identify and POS-tag English or English-transliterated words in each text. For all API calls, we set the temperature to 0 and top-$p$ to 1 to prioritize deterministic and accurate outputs over creative generation. Additionally, for tweets collected through keyword search on Twitter (X), where the presence of code-mixed words is guaranteed by the retrieval procedure, we rerun the annotation prompt up to three additional times in cases where the model returns no code-mixed words, which may occur due to occasional annotation failures. This prompting step inherently classifies the texts into \textit{code-mixed} and \textit{non-code-mixed}; therefore, we discard texts that contain no code-mixed words. We instruct the model to assign one of the 17 Universal Dependencies (UD) part-of-speech tags (\texttt{ADJ}, \texttt{ADP}, \texttt{ADV}, \texttt{AUX}, \texttt{CCONJ}, \texttt{DET}, \texttt{INTJ}, \texttt{NOUN}, \texttt{NUM}, \texttt{PART}, \texttt{PRON}, \texttt{PROPN}, \texttt{PUNCT}, \texttt{SCONJ}, \texttt{SYM}, \texttt{VERB}, and \texttt{X}) \cite{de2021universal} to each identified code-mixed word. The \texttt{X} tag is reserved for words whose part of speech cannot be determined or that do not fit any of the other categories, and we manually inspect all such cases. For English-transliterated words that are well-established and widely used loanwords in Persian, we instruct the model not to treat them as code-mixed. The exclusion of well-established loanwords is determined through a combination of a predefined list, LLM-based assessment, and human verification, with the final inclusion or exclusion decision based on their combined evidence. Mentions and hashtags are also excluded from the code-mixing analysis. Moreover, English-transliterated proper nouns, including names of people, diseases and medications, medical terms, sports teams, companies and brands, phone and car models, cities and countries, movies and books, and social media platforms (e.g., Twitter, Instagram, and YouTube), are likewise excluded from the code-mixing analysis.

Because each platform contains specific English keywords that are naturally and widely used by its users, such as \textit{timeline} and \textit{fave star} on Twitter, \textit{story} and \textit{live} on Instagram, and \textit{DigiPay} and \textit{DigiExpress} on Digikala, we design platform-specific prompts and instruct the LLM not to treat these items as code-mixed when they appear in Persian script. Appendix \ref{apx:prompts} provides the platform-specific POS annotation prompts, together with their English translations.

To evaluate the reliability of the automatic annotations, we randomly sample 150 texts from \textsc{PERCEPT}, containing 268 code-mixed words, with an equal number of texts (50) selected from each platform. Two native Persian-speaking annotators with prior NLP coursework independently verify the automatically identified code-mixed words and their UPOS tags following the Universal Dependencies guidelines. Annotators are instructed not to use AI assistance. Inter-annotator agreement yields a precision of 81.2\%, a recall of 91.5\%, and an F1 score of 86.0\% for code-mixed word identification. For POS tagging, the annotators achieve a Cohen's $\kappa$ of 0.894 with 93.2\% observed agreement, indicating almost perfect agreement. 

Disagreements are subsequently adjudicated by the first author to produce the final gold annotations. We then compare Gemini's annotations with the gold labels. For code-mixed word identification, Gemini achieves a precision of 95.5\%, a recall of 86.6\%, and an F1 score of 90.8\%. For POS tagging of correctly identified code-mixed words, Gemini achieves an accuracy of 97.0\% and a macro F1 score of 76.8\%. Considering the complete annotation pipeline, including both word identification and POS assignment, Gemini achieves an F1 score of 88.1\%. These results indicate that LLM-generated annotations closely align with human annotations, supporting the reliability of the automatic annotation process.

\subsection{Topic Detection}

To identify the topics of tweets and comments, we employ Gemini 3.5 Flash, following the same approach as in the previous section. We develop a topic taxonomy informed by existing topic-classification frameworks for media and social-media content \cite{martin2019we, iptc_media_topics}, adapting it to the characteristics of our dataset. The taxonomy comprises seven broad categories: Brand \& Business, Industry \& Commerce, Current Events \& News, Lifestyle \& Personal Interests, Arts, Entertainment \& Culture, Society \& Identity, and Platform-Specific \& Meta Topics. In each prompt, we provide the LLM with 50 texts, definitions of the seven categories, and several examples to facilitate consistent topic classification. Together, these categories cover a broad spectrum of topics discussed by Persian-speaking users on Twitter (X), Instagram, and Digikala. If a text cannot be confidently assigned to any predefined category, it is labeled as \textit{other}. This option is available to both the LLM and human annotators to avoid forcing an inappropriate topic assignment. Appendix~\ref{apx:topic} provides the prompt used for topic detection, together with its English translation.

To validate the automatic topic annotations, we use the same sample employed for the human validation of POS tagging. Two annotators independently annotate the topic of each text without access to the labels generated by Gemini, following the annotation guidelines provided in the prompt (see Appendix \ref{apx:annotation} for the complete annotation guidelines). Inter-annotator agreement achieves a Cohen's $\kappa$ of 0.738 with 79.3\% observed agreement, indicating substantial agreement between the annotators. 

After establishing the gold annotations, we evaluate the alignment between Gemini-generated topic labels and the human-annotated gold labels. Gemini achieves a Cohen's $\kappa$ of 0.808 and Krippendorff's $\alpha$ of 0.808, indicating substantial agreement with human judgments. The model obtains an observed agreement of 84.7\%, with a weighted F1-score of 84.4\% and a macro F1-score of 65.6\%. While the weighted F1 reflects the overall accuracy under the original topic distribution, the lower macro F1 indicates that performance varies across topics, particularly for less frequent categories.

\section{Main Results}

\subsection{POS Distribution}

Table~\ref{tab:posDist} presents the distribution of Universal POS tags among code-mixed words across the three platforms. Overall, \texttt{NOUN} is by far the most frequent POS category on all platforms, accounting for 60.4\% of code-mixed words in Twitter (X), 59.9\% in Instagram, and 45.4\% in Digikala. The remaining distributions differ across platforms: \texttt{VERB} is the second most frequent category in Twitter (X), whereas \texttt{ADJ} ranks second in Instagram and \texttt{PROPN} in Digikala. These findings indicate that code-mixing predominantly occurs in nouns, while the secondary patterns appear to reflect platform-specific language use. This observation is consistent with prior studies of English code-mixing in other languages, which likewise report the predominance of nouns and adjectives among POS categories in social media text \cite{farooq2026morphosyntactic,dongen2017analysis}. Additionally, the unusually high proportion of \texttt{PROPN} in Digikala is likely attributable to the frequent use of English brand names and product model names in e-commerce listings, a pattern that has also been observed in advertising in other languages \cite{jia2008glocalization}.

\begin{table}[htbp]
\centering
\small
\setlength{\tabcolsep}{3pt}
\begin{tabular}{lrrr}
\toprule
\textbf{POS} & \textbf{Twitter (X)} & \textbf{Instagram} & \textbf{Digikala} \\
\midrule
\texttt{ADJ}   & \textbf{723} (12.8\%)  & \textbf{316} (19.3\%) & \textbf{315} (9.0\%) \\
\texttt{ADP}   & 17 (0.3\%)    & 4 (0.2\%)    & 13 (0.4\%) \\
\texttt{ADV}   & 65 (1.1\%)    & 11 (0.7\%)   & 5 (0.1\%) \\
\texttt{AUX}   & 5 (0.1\%)     & 0 (0.0\%)    & 0 (0.0\%) \\
\texttt{CCONJ} & 3 (0.1\%)     & 0 (0.0\%)    & 2 (0.1\%) \\
\texttt{DET}   & 14 (0.2\%)    & 0 (0.0\%)    & 4 (0.1\%) \\
\texttt{INTJ}  & 25 (0.4\%)    & 16 (1.0\%)   & 0 (0.0\%) \\
\texttt{NOUN}  & \textbf{3,419} (60.4\%) & \textbf{979} (59.9\%) & \textbf{1,583} (45.4\%) \\
\texttt{NUM}   & 10 (0.2\%)    & 4 (0.2\%)    & 88 (2.5\%) \\
\texttt{PART}  & 2 (0.0\%)     & 0 (0.0\%)    & 0 (0.0\%) \\
\texttt{PRON}  & 15 (0.3\%)    & 1 (0.1\%)    & 0 (0.0\%) \\
\texttt{PROPN} & 58 (1.0\%)    & 11 (0.7\%)   & \textbf{1,220} (35.0\%) \\
\texttt{PUNCT} & 0 (0.0\%)     & 0 (0.0\%)    & 0 (0.0\%) \\
\texttt{SCONJ} & 1 (0.0\%)     & 0 (0.0\%)    & 0 (0.0\%) \\
\texttt{SYM}   & 0 (0.0\%)     & 0 (0.0\%)    & 5 (0.1\%) \\
\texttt{VERB}  & \textbf{1,302} (23.0\%) & \textbf{289} (17.7\%) & 252 (7.2\%) \\
\texttt{X}     & 1 (0.0\%)     & 4 (0.2\%)    & 3 (0.1\%) \\
\midrule
\textbf{Overall} & {5,660} & {1,635} & {3,490} \\
\bottomrule
\end{tabular}
\caption{Distribution of code-mixed words across POS categories by platform. Each cell reports the number of words and the percentage of code-mixed words within the corresponding platform. POS categories with fewer than 30 code-mixed words in a platform should be interpreted with caution due to their small sample sizes.}
\label{tab:posDist}
\end{table}

\subsection{Mixing Position}

To analyze the position of code-mixing within utterances, we compute the character-level start offset of each code-mixed word in the original text. Each occurrence is assigned a normalized position by dividing its start offset by the total character length of the utterance. We then categorize occurrences into initial ($\leq 33\%$), medial ($33$--$66\%$), and final ($>66\%$) positions.

Table~\ref{tab:positionDist} presents the distribution of code-mixed word positions across the three platforms. The positional distributions are highly consistent, with approximately one-third of code-mixed words occurring in the initial and medial positions and slightly fewer in the final position. To assess whether these differences are statistically significant, we perform a chi-square test of independence, which finds no significant association between platform and code-mixed word position ($\chi^2(4)=8.38$, $p=0.079$). This result further supports the observed consistency of positional distributions across platforms.

\begin{table}[htbp]
\centering
\small
\setlength{\tabcolsep}{3pt}
\begin{tabular}{lrrr}
\toprule
\textbf{Position} & \textbf{Twitter (X)} & \textbf{Instagram} & \textbf{Digikala} \\
\midrule
Initial     & 2,106 (37.2\%)    & 571 (34.9\%)    & 1,222 (35.0\%) \\
Medial   & 2,019 (35.7\%)    & 587 (35.9\%)    & 1,242 (35.6\%) \\
Final      & 1,535 (27.1\%)    & 477 (29.2\%)    & 1,026 (29.4\%) \\
\midrule
\textbf{Overall} & {5,660} & {1,635} & {3,490} \\
\bottomrule
\end{tabular}
\caption{Distribution of code-mixed words across positional categories per platform. Each cell shows count / percentage. Percentages are computed relative to the total number of code-mixed words in that platform (or overall).}
\label{tab:positionDist}
\end{table}

\subsection{Triggering Effect}

To operationalize the triggering effect at the utterance level, we measure the proportion of utterances containing more than one code-mixed word. Specifically, we compute the proportion of utterances with multiple code-mixed words for each platform. Since utterance length can affect the likelihood of observing multiple code-mixed words, we additionally control for length differences across platforms by computing the triggering rate within utterance-length groups and averaging across groups. Table~\ref{tab:triggering} presents the length-controlled number and percentage of utterances exhibiting the triggering effect for each platform and overall.

Digikala exhibits the highest triggering effect (42.0\%), indicating that code-mixed words are more likely to co-occur within the same utterance on this platform. This pattern may be attributable to the nature of e-commerce platforms, where users frequently mention multiple product features, brand names, or technical specifications within a single review. Twitter (X) ranks second, with 27\% of utterances containing multiple code-mixed words, while Instagram shows the lowest proportion (10.2\%).

\begin{table}[htbp]
\centering
\small
\begin{tabular}{lc}
\toprule
\textbf{Platform} & \textbf{Triggering Effect} \\
\midrule
Twitter (X) & 1,139 (27.1\%) \\
Instagram & 148 (10.2\%) \\
Digikala    & 747 (42.0\%) \\
\bottomrule
\end{tabular}
\caption{Length-controlled triggering effect of code-mixing across platforms. Each cell shows the number of utterances with more than one code-mixed word and the corresponding adjusted percentage.}
\label{tab:triggering}
\end{table}

\subsection{Topic Variation}

\begin{figure}[htbp] 
    \centering
    \includegraphics[width=1\columnwidth]{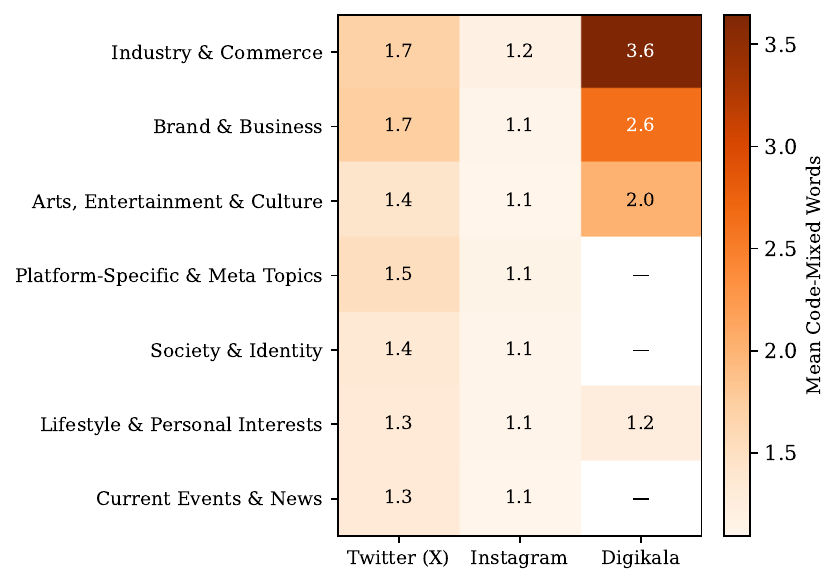} 
    \caption{Heatmap of the degree of code-mixing across discussion topics and platforms, measured as the mean number of code-mixed words per utterance. Darker cells indicate a higher degree of code-mixing, while white cells denote the absence of data for the corresponding topic--platform combination.}
    \label{fig:heatmapTopic}
\end{figure}

To assess how the degree of code-mixing varies across discussion topics, we calculate the mean number of code-mixed words per utterance for each platform--topic combination. Figure~\ref{fig:heatmapTopic} presents these values. The number of utterances underlying each platform--topic combination is reported in Appendix~\ref{apx:topicCounts}. Across all three platforms, \textit{Industry \& Commerce} and \textit{Brand \& Business} exhibit the highest mean number of code-mixed words per utterance, indicating that discussions related to commercial activities and brands consistently involve a greater degree of code-mixing than other topics. One possible explanation is that many technology- and commerce-related concepts are commonly referred to by their English names in informal Persian, even when Persian equivalents exist \cite{sahranavard2025english}. 

While these two topics consistently rank highest, the influence of topic varies substantially across platforms. On Instagram, the mean number of code-mixed words remains nearly constant across all topics (1.1--1.2), suggesting that once users code-mix, they typically insert only a single English word regardless of the discussion topic. In contrast, Twitter (X) exhibits moderate variation across topics, whereas Digikala shows the strongest topic dependence, with commerce-related discussions containing substantially more code-mixed words than other topics. This pattern is expected, as Digikala discussions are primarily centered on products and commercial activities, while Instagram encompasses a much broader range of discussion topics.

\section{Discussion and Conclusion}

In this work, we present \textsc{PERCEPT}, a dataset of 6,800 Persian--English code-mixed tweets and comments collected from Twitter (X), Instagram, and Digikala. Using Gemini 3.5 Flash, we automatically identify code-mixed words, assign Universal Dependencies POS tags, and detect the topic of each text. Human validation demonstrates high agreement between the LLM-generated annotations and gold annotations, confirming the reliability of the dataset. Our analyses show that code-mixing occurs predominantly in nouns, while secondary POS distributions vary across platforms, with \texttt{VERB} ranking second in Twitter (X), \texttt{ADJ} in Instagram, and \texttt{PROPN} in Digikala. We further find that the positional distribution of code-mixed words is remarkably consistent across platforms, with most occurrences appearing in the initial and medial portions of an utterance. In contrast, the triggering effect exhibits substantial platform-specific variation and is most pronounced in Digikala, likely reflecting the e-commerce nature of the platform. Finally, topic analysis reveals that \textit{Industry \& Commerce} and \textit{Brand \& Business} contain the highest degree of code-mixing across the analyzed platforms.

The annotation framework proposed in this work is readily transferable to other low-resource languages, many of which similarly lack large-scale POS-annotated code-mixed corpora. Since the prompting strategy is largely language-independent, researchers can adapt it by translating the prompts and applying them to conversational data collected from social media or other online platforms. Beyond cross-lingual applications, \textsc{PERCEPT} provides a valuable resource for Persian NLP. The dataset can support the development and evaluation of POS taggers and other code-mixed NLP systems, including language identification, named entity recognition, syntactic parsing, sentiment analysis, and text generation. It also enables future linguistic studies on the structural, lexical, and sociolinguistic properties of Persian--English code-mixing.

\section*{Limitations}

This study has several limitations. First, the linguistic analyses presented in this work are specific to Persian--English code-mixed text and may not directly generalize to other language pairs with different linguistic and sociocultural characteristics. Nevertheless, the proposed annotation framework is based on the language-independent Universal Dependencies tagset and can be readily adapted to other code-mixed languages, enabling future comparative studies across diverse language pairs.

Second, \textsc{PERCEPT} is constructed from three platforms: Twitter (X), Instagram, and Digikala, which capture different types of social media and e-commerce interactions but do not represent the full spectrum of online communication. Future work could extend the dataset to additional platforms to provide a more comprehensive understanding of Persian--English code-mixing.

A further limitation concerns the construction of the Twitter (X) subset. In addition to organically collected posts, we include posts retrieved through targeted searches for frequent code-mixed words identified in the initial sample. Although this procedure increases the coverage of code-mixed content, it may introduce selection bias by overrepresenting the syntactic patterns and topics associated with these frequent words. Consequently, the X subset may not fully reflect the natural distribution of Persian--English code-mixing on the platform, and cross-platform comparisons should be interpreted with this sampling difference in mind.

Finally, this work focuses on POS tagging of code-mixed words and the linguistic analyses enabled by these annotations. Other annotation layers, such as named entity recognition, dependency parsing, token-level language identification, matrix language identification, and additional downstream NLP tasks, remain unexplored and constitute promising directions for future research.

Despite these limitations, we publicly release \textsc{PERCEPT}, the first large-scale POS-annotated Persian--English code-mixed corpus, together with the analysis code and the annotation prompts used in this work. We hope this resource will facilitate future research on Persian code-mixing and support the development of multilingual NLP systems for low-resource languages.

\bibliography{custom}

@inproceedings{sabri2021sentiment,
  title={Sentiment analysis of persian-english code-mixed texts},
  author={Sabri, Nazanin and Edalat, Ali and Bahrak, Behnam},
  booktitle={2021 26th International Computer Conference, Computer Society of Iran (CSICC)},
  pages={1--4},
  year={2021},
  organization={IEEE}
}

@inproceedings{chakravarthi2020sentiment,
  title={A sentiment analysis dataset for code-mixed Malayalam-English},
  author={Chakravarthi, Bharathi Raja and Jose, Navya and Suryawanshi, Shardul and Sherly, Elizabeth and McCrae, John Philip},
  booktitle={Proceedings of the 1st Joint workshop on spoken language technologies for under-resourced languages (SLTU) and collaboration and computing for under-resourced languages (CCURL)},
  pages={177--184},
  year={2020}
}

@inproceedings{alam2025bnsentmix,
  title={BnSentMix: A diverse Bengali-English code-mixed dataset for sentiment analysis},
  author={Alam, Sadia and Ishmam, Md Farhan and Alvee, Navid Hasin and Siddique, Md Shahnewaz and Hossain, Md Azam and Kamal, Abu Raihan Mostofa},
  booktitle={Proceedings of the First Workshop on Language Models for Low-Resource Languages},
  pages={68--77},
  year={2025}
}

@inproceedings{uthpala2024sinhala,
  title={Sinhala-English code-mixed language dataset with sentiment annotation},
  author={Uthpala, DK and Thirukumaran, S},
  booktitle={2024 4th International Conference on Advanced Research in Computing (ICARC)},
  pages={184--188},
  year={2024},
  organization={IEEE}
}

@inproceedings{vyas2014pos,
  title={Pos tagging of english-hindi code-mixed social media content},
  author={Vyas, Yogarshi and Gella, Spandana and Sharma, Jatin and Bali, Kalika and Choudhury, Monojit},
  booktitle={Proceedings of the 2014 conference on empirical methods in natural language processing (EMNLP)},
  pages={974--979},
  year={2014}
}

@inproceedings{singh2018twitter,
  title={A Twitter corpus for Hindi-English code mixed POS tagging},
  author={Singh, Kushagra and Sen, Indira and Kumaraguru, Ponnurangam},
  booktitle={Proceedings of the sixth international workshop on natural language processing for social media},
  pages={12--17},
  year={2018}
}

@inproceedings{barman2016part,
  title={Part-of-speech tagging of code-mixed social media content: Pipeline, stacking and joint modelling},
  author={Barman, Utsab and Wagner, Joachim and Foster, Jennifer},
  booktitle={Proceedings of the second workshop on computational approaches to code switching},
  pages={30--39},
  year={2016}
}

@inproceedings{golazizian2020irony,
  title={Irony detection in Persian language: A transfer learning approach using emoji prediction},
  author={Golazizian, Preni and Sabeti, Behnam and Asli, Seyed Arad Ashrafi and Majdabadi, Zahra and Momenzadeh, Omid and Fahmi, Reza},
  booktitle={Proceedings of the Twelfth Language Resources and Evaluation Conference},
  pages={2839--2845},
  year={2020}
}

@inproceedings{sabri2021emopars,
  title={Emopars: A collection of 30k emotion-annotated persian social media texts},
  author={Sabri, Nazanin and Akhavan, Reyhane and Bahrak, Behnam},
  booktitle={Proceedings of the student research workshop associated with RANLP 2021},
  pages={167--173},
  year={2021}
}

@article{kebriaei2024persian,
  title={Persian offensive language detection},
  author={Kebriaei, Emad and Homayouni, Ali and Faraji, Roghayeh and Razavi, Armita and Shakery, Azadeh and Faili, Heshaam and Yaghoobzadeh, Yadollah},
  journal={Machine Learning},
  volume={113},
  number={7},
  pages={4359--4379},
  year={2024},
  publisher={Springer}
}

@article{elahimanesh2025emotion,
  title={Emotion Alignment: Discovering the Gap Between Social Media and Real-World Sentiments in Persian Tweets and Images},
  author={Elahimanesh, Sina and Mohammadkhani, Mohammadali and Kasaei, Shohreh},
  journal={arXiv preprint arXiv:2504.10662},
  year={2025}
}

@inproceedings{heidari2020producing,
  title={Producing an instagram dataset for persian language sentiment analysis using crowdsourcing method},
  author={Heidari, Mahsa and Shamsinejad, Pirooz},
  booktitle={2020 6th International Conference on Web Research (ICWR)},
  pages={284--287},
  year={2020},
  organization={IEEE}
}

@inproceedings{nazarizadeh2025parsoffensive,
  title={Parsoffensive: Persian Offensive Comments Dataset},
  author={Nazarizadeh, Ali and Sayyadpour, Minoo and Ebrahimkhani, Omid and Iravani, Amirmasoud and Shirazi, Golnaz Ghannadan and Behravan, Mojgan},
  booktitle={2025 11th International Conference on Web Research (ICWR)},
  pages={100--104},
  year={2025},
  organization={IEEE}
}

@inproceedings{asli2020optimizing,
  title={Optimizing annotation effort using active learning strategies: A sentiment analysis case study in persian},
  author={Asli, Seyed Arad Ashrafi and Sabeti, Behnam and Majdabadi, Zahra and Golazizian, Preni and Fahmi, Reza and Momenzadeh, Omid},
  booktitle={Proceedings of the Twelfth Language Resources and Evaluation Conference},
  pages={2855--2861},
  year={2020}
}

@article{ghafouri2025pinlid,
  title={PinLID: a dataset for Pinglish language identiftcation based on code-mixing sentence on unstructured resources},
  author={Ghafouri, Arash and Naderi, Hasan and Firouzmandi, Mahdi},
  journal={Language Resources and Evaluation},
  volume={59},
  number={3},
  pages={3215--3241},
  year={2025},
  publisher={Springer}
}

@misc{google2026gemini35,
  author       = {{Google}},
  title        = {What's New in Gemini 3.5 Flash},
  year         = {2026},
  howpublished = {\url{https://ai.google.dev/gemini-api/docs/whats-new-gemini-3.5}},
  note         = {Google AI for Developers. Accessed: 2026-07-07}
}

@article{de2021universal,
  title={Universal dependencies},
  author={De Marneffe, Marie-Catherine and Manning, Christopher D and Nivre, Joakim and Zeman, Daniel},
  journal={Computational linguistics},
  volume={47},
  number={2},
  pages={255--308},
  year={2021}
}

@article{farooq2026morphosyntactic,
  title={MORPHOSYNTACTIC ANALYSIS OF URDU--ENGLISH CODE-SWITCHING OF TWEETER ACTIVISTS IN PAKISTAN},
  author={Farooq, Mahwish and Eesa, Muhammad and Sahar, Abeera and Farooq, Sahirish and Babu, Sameer and Khan, Asma and Hassan, Roshidah},
  journal={Veredas do Direito},
  volume={23},
  number={8},
  pages={e235012--e235012},
  year={2026}
}

@article{dongen2017analysis,
  title={Analysis and prediction of Dutch-English code-switching in Dutch social media messages},
  author={Dongen, Nina},
  journal={Masterscriptie. Universiteit van Amsterdam},
  year={2017}
}

@article{jia2008glocalization,
  title={Glocalization and English mixing in advertising in Taiwan: Its discourse domains, linguistic patterns, cultural constraints, localized creativity, and socio-psychological effects},
  author={Jia-Ling, Hsu},
  journal={Journal of Creative Communications},
  volume={3},
  number={2},
  pages={155--183},
  year={2008},
  publisher={SAGE Publications Sage India: New Delhi, India}
}

@article{sahranavard2025english,
  title={English Loanwords in Persian},
  author={Sahranavard, Neda},
  journal={Iranian Studies},
  volume={58},
  number={3},
  pages={595--610},
  year={2025},
  publisher={Cambridge University Press}
}

@article{nordin2023code,
  title={Code-switching and code-mixing among users of social media},
  author={Nordin, Nur Rasyidah Mohd},
  journal={Jurnal Javanologi},
  volume={6},
  number={2},
  pages={1267--1273},
  year={2023}
}

@phdthesis{barman2019automatic,
  title={Automatic processing of code-mixed social media content},
  author={Barman, Utsab},
  year={2019}
}

@inproceedings{singh2018automatic,
  title={Automatic normalization of word variations in code-mixed social media text},
  author={Singh, Rajat and Choudhary, Nurendra and Shrivastava, Manish},
  booktitle={International Conference on Computational Linguistics and Intelligent Text Processing},
  pages={371--381},
  year={2018},
  organization={Springer}
}

@inproceedings{sheth2025comi,
    title = "{COMI}-{LINGUA}: Expert Annotated Large-Scale Dataset for Multitask {NLP} in {H}indi-{E}nglish Code-Mixing",
    author = "Sheth, Rajvee  and Beniwal, Himanshu  and Singh, Mayank",
    booktitle = "Findings of the Association for Computational Linguistics: EMNLP 2025",
    month = nov,
    year = "2025",
    address = "Suzhou, China",
    publisher = "Association for Computational Linguistics",
    pages = "7973--7992"
}

@article{moradi2019structural,
    author = {Moradi, Hamzeh and Chen, Jianbo},
    title = {Structural Analysis of {Persian-English} Reverse Code-Switching and Code-Mixing},
    journal = {Vestnik Volgogradskogo gosudarstvennogo universiteta. Seriya 2, Yazykoznanie [Science Journal of Volgograd State University. Linguistics]},
    year = {2019},
    volume = {18},
    number = {1},
    pages = {122--131},
    doi = {10.15688/jvolsu2.2019.1.10}
}

@inproceedings{davardoust2024dark,
    author = {Davardoust, Hadi and Zare, Hadi and RafieeZade, Hossein},
    title = {The dark side of Instagram: A large dataset for identifying Persian harmful comments},
    booktitle = {SoCal NLP Symposium 2024},
    year = {2024}
}

@inproceedings{kumar2018aggression,
    title = "Aggression-annotated Corpus of {H}indi-{E}nglish Code-mixed Data",
    author = "Kumar, Ritesh  and Reganti, Aishwarya N.  and Bhatia, Akshit  and Maheshwari, Tushar",
    booktitle = "Proceedings of the Eleventh International Conference on Language Resources and Evaluation ({LREC} 2018)",
    month = may,
    year = "2018",
    address = "Miyazaki, Japan",
    publisher = "European Language Resources Association (ELRA)",
}

@misc{iptc_media_topics,
  author       = {{International Press Telecommunications Council (IPTC)}},
  title        = {{IPTC Media Topics}},
  howpublished = {\url{https://iptc.org/standards/media-topics/}},
  note         = {Accessed: 2026-10-09}
}

@incollection{martin2019we,
  title={What We Share: Genre and Topicality on Facebook and Twitter},
  author={Martin, Fiona},
  booktitle={Sharing News Online: Commendary Cultures and Social Media News Ecologies},
  pages={129--156},
  year={2019},
  publisher={Springer}
}

@inproceedings{solorio2008part,
  title={Part-of-speech tagging for English-Spanish code-switched text},
  author={Solorio, Thamar and Liu, Yang},
  booktitle={Proceedings of the 2008 Conference on Empirical Methods in Natural Language Processing},
  pages={1051--1060},
  year={2008}
}

@inproceedings{jamatia2016collecting,
  title={Collecting and annotating indian social media code-mixed corpora},
  author={Jamatia, Anupam and Gamb{\"a}ck, Bj{\"o}rn and Das, Amitava},
  booktitle={International Conference on Intelligent Text Processing and Computational Linguistics},
  pages={406--417},
  year={2016},
  organization={Springer}
}

@inproceedings{bhat2018universal,
  title={Universal Dependency parsing for Hindi-English code-switching},
  author={Bhat, Irshad and Bhat, Riyaz Ahmad and Shrivastava, Manish and Sharma, Dipti Misra},
  booktitle={Proceedings of the 2018 Conference of the North American Chapter of the Association for Computational Linguistics: Human Language Technologies, Volume 1 (Long Papers)},
  pages={987--998},
  year={2018}
}

@article{ccetinouglu2023two,
  title={Two languages, one treebank: building a Turkish--German code-switching treebank and its challenges},
  author={{\c{C}}etino{\u{g}}lu, {\"O}zlem and {\c{C}}{\"o}ltekin, {\c{C}}a{\u{g}}r{\i}},
  journal={Language Resources and Evaluation},
  volume={57},
  number={2},
  pages={545--579},
  year={2023},
  publisher={Springer}
}

\appendix

\section{Example Prompts}\label{apx:prompts}

\subsection{Prompts for POS Annotation}

\begin{figure*}[!t]
  \centering
  \includegraphics[width=\textwidth]{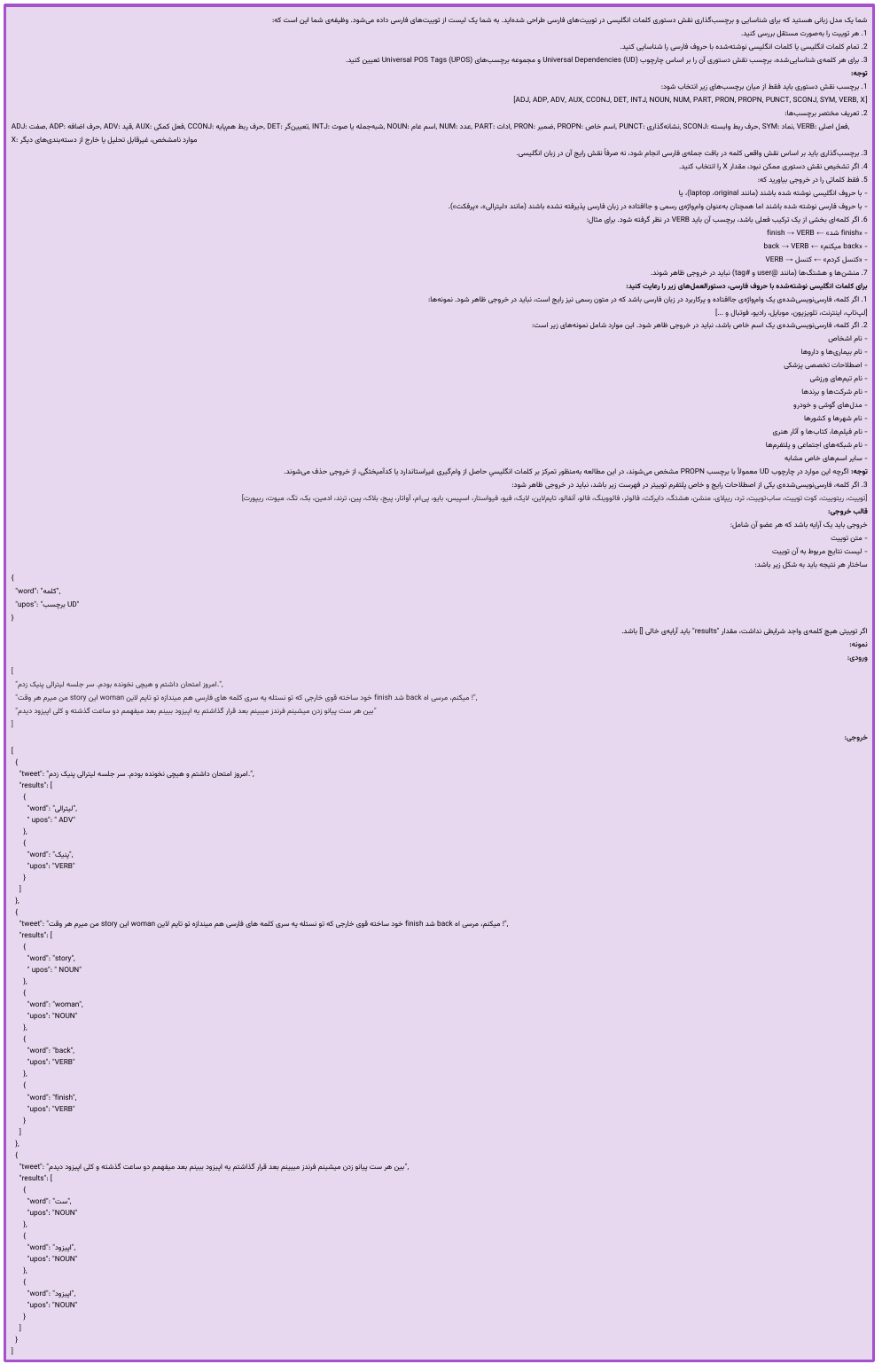}
  \caption{Example prompt for identifying and annotating the parts of speech of English or English-transliterated words in Persian tweets on Twitter (X).}
  \label{fig:twitterPrompt}
\end{figure*}

\begin{figure*}[!t]
  \centering
  \includegraphics[width=\textwidth]{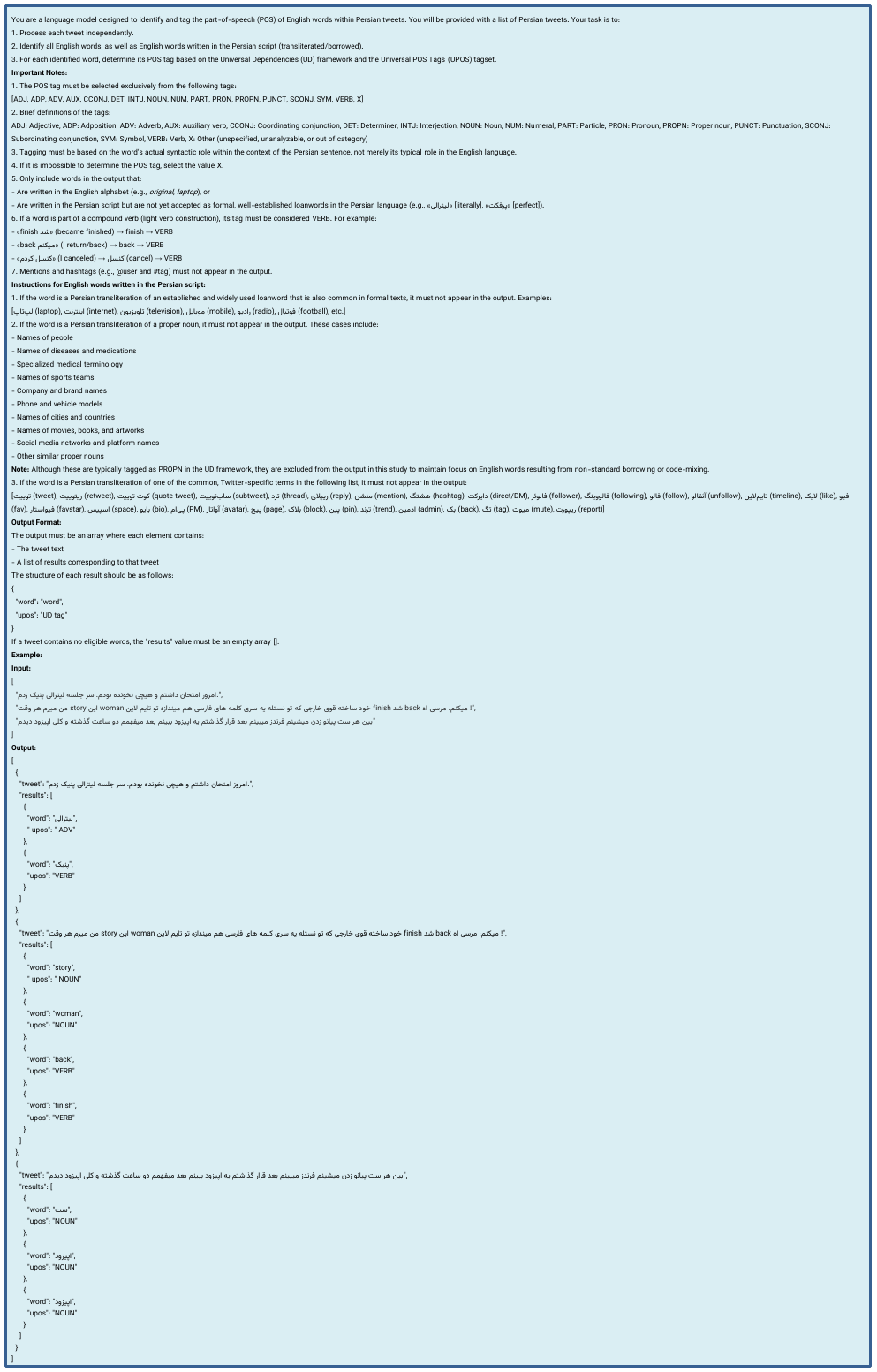}
  \caption{English translation of the example prompt for identifying and annotating the parts of speech of English or English-transliterated words in Persian tweets on Twitter (X).}
  \label{fig:twitterPromptEn}
\end{figure*}

\begin{figure*}[!t]
  \centering
  \includegraphics[width=\textwidth]{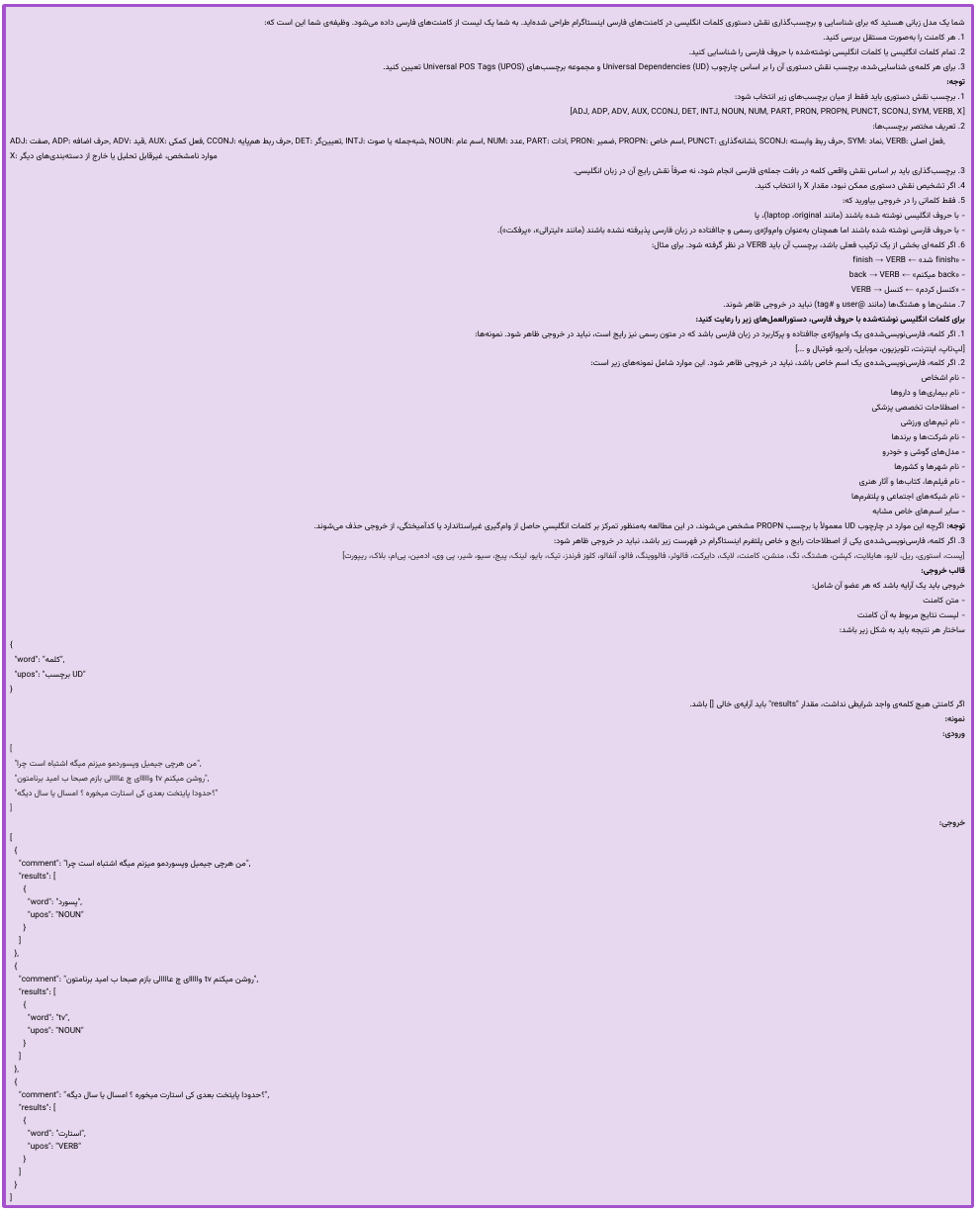}
  \caption{Example prompt for identifying and annotating the parts of speech of English or English-transliterated words in Persian comments on Instagram.}
  \label{fig:instaPrompt}
\end{figure*}

\begin{figure*}[!t]
  \centering
  \includegraphics[width=\textwidth]{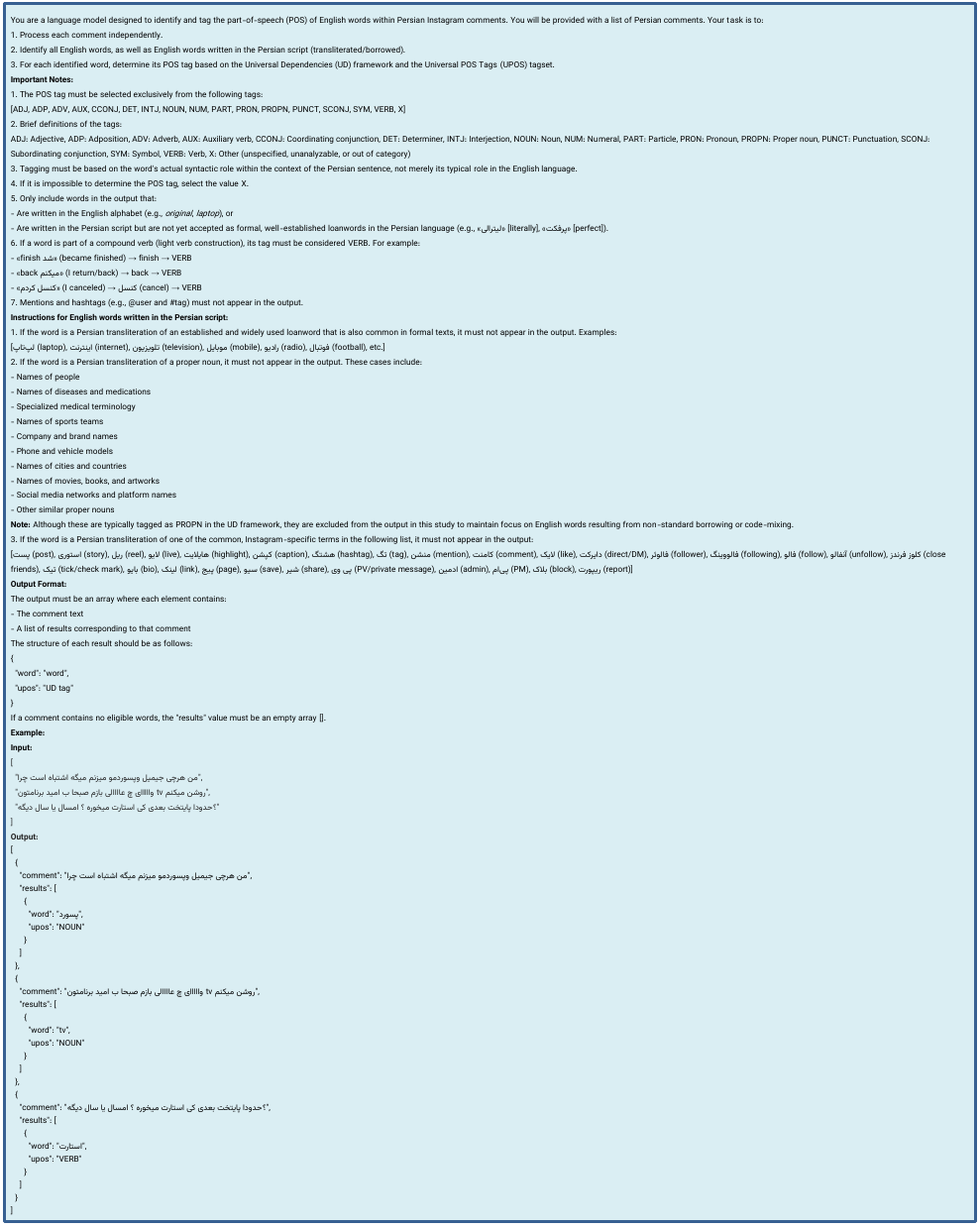}
  \caption{English translation of the example prompt for identifying and annotating the parts of speech of English or English-transliterated words in Persian comments on Instagram.}
  \label{fig:instaPromptEn}
\end{figure*}

\begin{figure*}[!t]
  \centering
  \includegraphics[width=\textwidth]{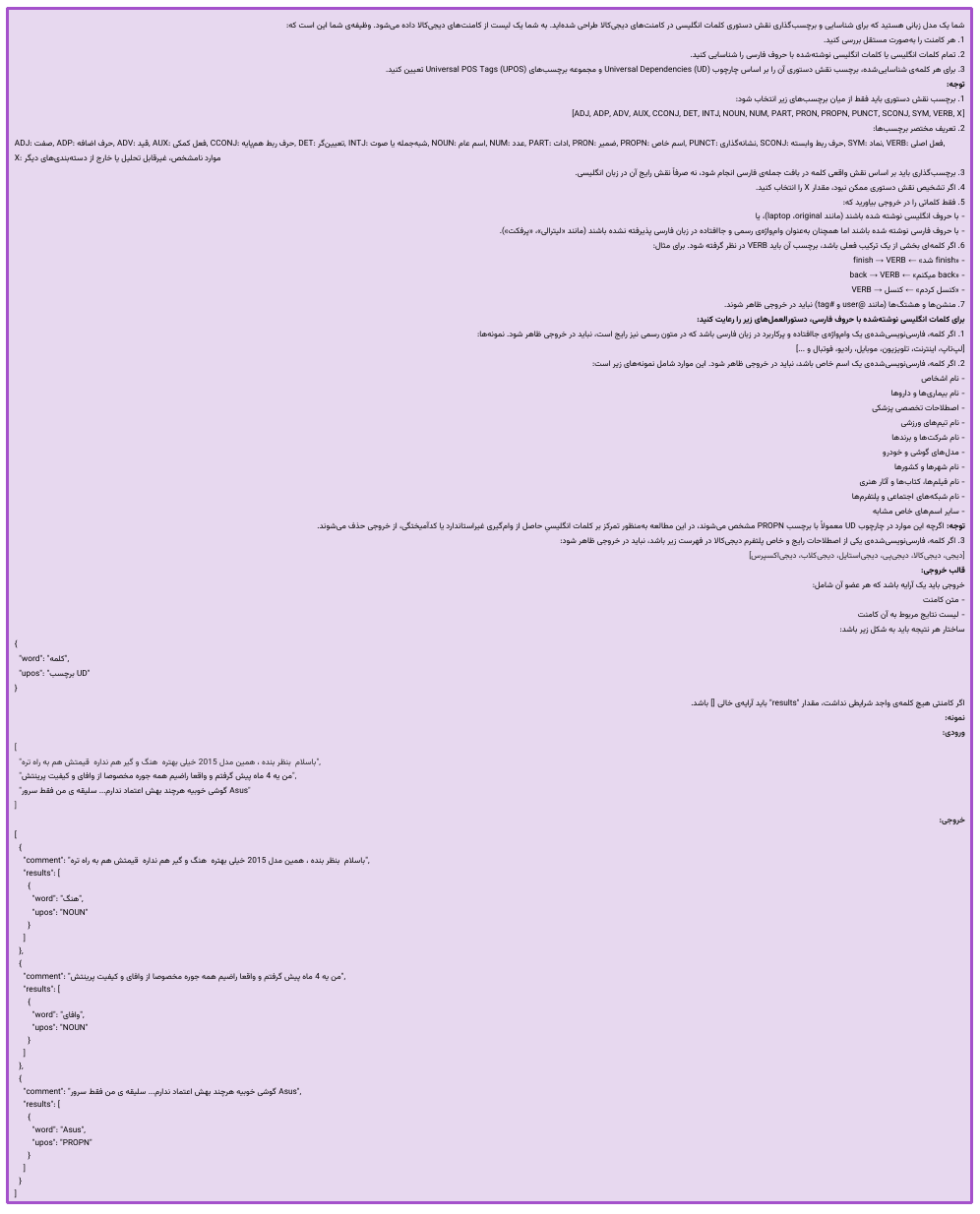}
  \caption{Example prompt for identifying and annotating the parts of speech of English or English-transliterated words in Persian comments on Digikala.}
  \label{fig:digikalaPrompt}
\end{figure*}

\begin{figure*}[!t]
  \centering
  \includegraphics[width=\textwidth]{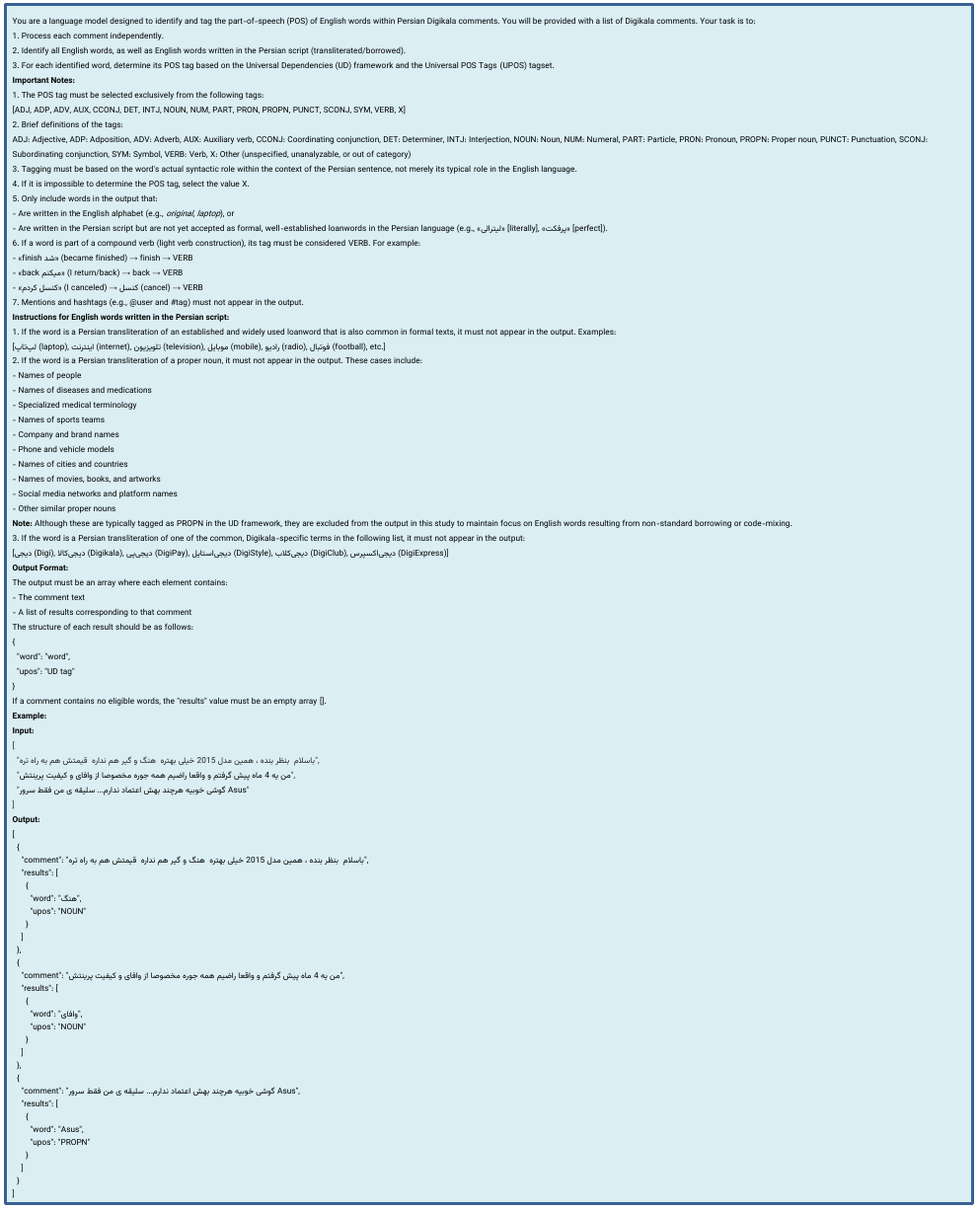}
  \caption{English translation of the example prompt for identifying and annotating the parts of speech of English or English-transliterated words in Persian comments on Digikala.}
  \label{fig:digikalaPromptEn}
\end{figure*}

Example prompts used to annotate the parts of speech of English or English-transliterated words in Persian tweets and comments across different platforms, along with their English translations, are provided in Figures \ref{fig:twitterPrompt}--\ref{fig:digikalaPromptEn}.

\subsection{Topic Detection}\label{apx:topic}

\begin{figure*}[!t]
  \centering
  \includegraphics[width=\textwidth]{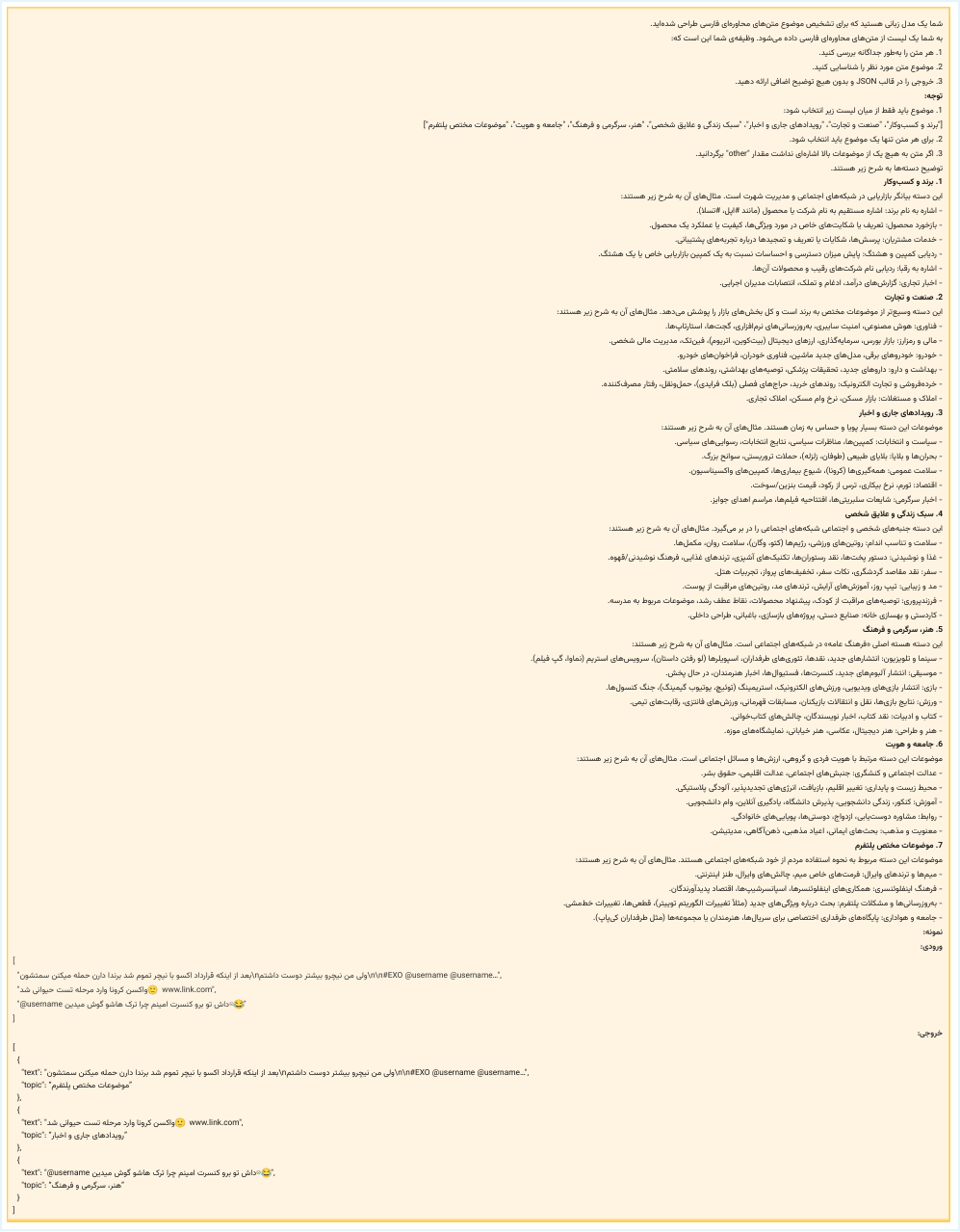}
  \caption{Example prompt for detecting the topics of Persian--English code-mixed tweets and comments.}
  \label{fig:topicPrompt}
\end{figure*}

\begin{figure*}[!t]
  \centering
  \includegraphics[width=\textwidth]{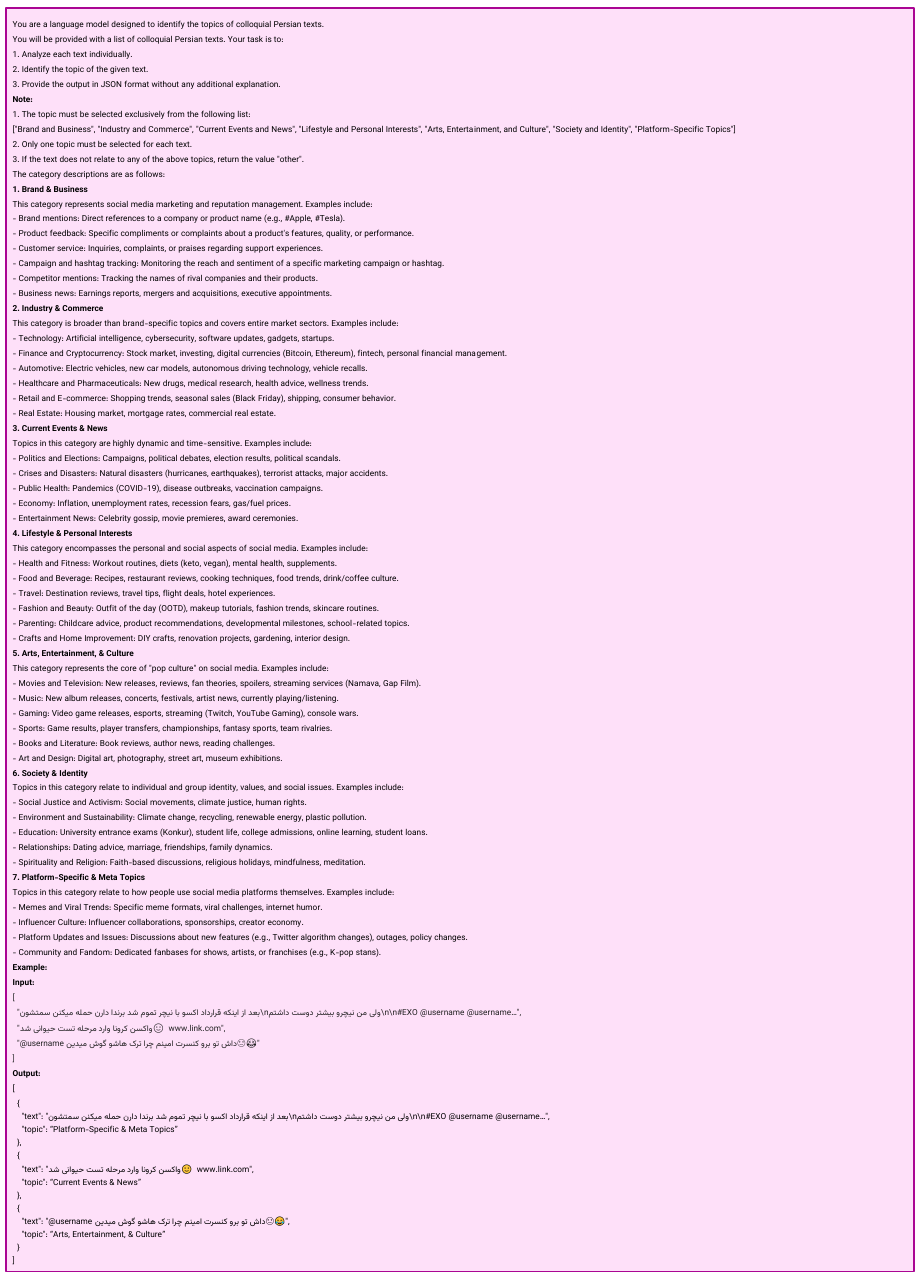}
  \caption{English translation of the example prompt for detecting the topics of Persian--English code-mixed tweets and comments.}
  \label{fig:topicPromptEn}
\end{figure*}

Figures \ref{fig:topicPrompt} and \ref{fig:topicPromptEn} illustrate the example prompt used to identify the topics of Persian--English code-mixed tweets and comments, along with its English translation.

\section{Human Annotation Protocol}\label{apx:annotation}

\subsection{Annotators}

\begin{itemize}
\item Number of annotators: 2
\item Native language: Persian
\item Education: undergraduate students in Computer Engineering
\item Prior annotation experience: basic NLP coursework
\item Compensation: course-related research assistance
\end{itemize}

\subsection{Annotation Task}

For each text, annotators are asked to:

\begin{enumerate}
\item Identify all Persian--English code-mixed words.
\item Assign a Universal Dependencies UPOS tag to each identified word.
\item Assign one topic category from the predefined taxonomy.
\end{enumerate}

\subsection{Key Instructions}

\begin{itemize}
\item Treat English words written in either Latin script or Persian transliteration as code-mixed if they are used as English lexical items.
\item Do not mark platform-specific product or interface names (e.g., DigiPay, DigiExpress, Instagram Story) as code-mixed when they function as proper names.
\item In multi-word English expressions, annotate each word separately with its own UPOS tag.
\item Follow the Universal Dependencies v2 definitions for POS tagging.
\item Do not use external AI tools during annotation.
\end{itemize}

\section{Number of Utterances by Platform and Topic}\label{apx:topicCounts}

\begin{table}[htbp]
\centering
\scriptsize
\setlength{\tabcolsep}{3pt}
\begin{tabular}{lrrr}
\toprule
\textbf{English Topic} & \textbf{Twitter (X)} & \textbf{Instagram} & \textbf{Digikala} \\
\midrule
Industry \& Commerce            & 344  & 78  & 28 \\
Brand \& Business               & 101  & 94  & 1284 \\
Arts, Entertainment \& Culture  & 553  & 491 & 2 \\
Platform-Specific \& Meta Topics & 674  & 211 & -- \\
Society \& Identity              & 1013 & 205 & -- \\
Lifestyle \& Personal Interests  & 500  & 105 & 4 \\
Current Events \& News           & 504  & 181 & -- \\
\bottomrule
\end{tabular}
\caption{Number of utterances underlying the means reported in Figure~\ref{fig:heatmapTopic}. Values denote the number of utterances for each platform--topic combination; -- indicates that no utterances are available for the corresponding combination.}
\label{tab:values}
\end{table}

\end{document}